%% file: root.tex
\documentclass[letterpaper, 10 pt, conference]{ieeeconf}
\usepackage{cite}
\usepackage[font=small]{caption}
\usepackage{hyperref}
\usepackage{cite}
\usepackage{dsfont}
\usepackage{amsmath,amssymb,amsfonts}
\usepackage{graphicx}
\usepackage{textcomp}
\usepackage{xcolor}
\usepackage{subcaption}
\usepackage{algorithm}
\usepackage{algpseudocode}
\usepackage{hyperref}
\usepackage[font=small]{caption} 

\def\BibTeX{{\rm B\kern-.05em{\sc i\kern-.025em b}\kern-.08em
    T\kern-.1667em\lower.7ex\hbox{E}\kern-.125emX}}

\title{Reward Learning from Suboptimal Demonstrations \\ with Applications in Surgical Electrocautery}
\author{Zohre Karimi$^*$, Shing-Hei Ho$^*$, Bao Thach, Alan Kuntz, Daniel S. Brown%
\thanks{* Equal Contribution. Robotics Center and the Kahlert School of Computing at the University of Utah, Salt Lake City, UT 84112, USA; (email: \{zohre.karimi, shinghei.ho, bao.thach, alan.kuntz, daniel.brown\}@utah.edu). 
This material is based upon work supported in part by the National Science Foundation under grant number 2133027. Any opinions, findings, and conclusions or recommendations expressed in this material are those of the authors and do not necessarily reflect the views of the NSF.}}
\date{September 2023}

\IEEEoverridecommandlockouts

\begin{document}

\maketitle
\begin{abstract}
    Automating robotic surgery via learning from demonstration (LfD) techniques is extremely challenging. This is because surgical tasks often involve sequential decision-making processes with complex interactions of physical objects and have low tolerance for mistakes. Prior works assume that all demonstrations are fully observable and optimal, which might not be practical in the real world. This paper introduces a sample-efficient method that learns a robust reward function from a limited amount of ranked suboptimal demonstrations consisting of partial-view point cloud observations. The method then learns a policy by optimizing the learned reward function using reinforcement learning (RL). We show that using a learned reward function to obtain a policy is more robust than pure imitation learning. We apply our approach on a physical surgical electrocautery task and demonstrate that our method can perform well even when the provided demonstrations are suboptimal and the observations are high-dimensional point clouds. Code and videos available here: \href{https://sites.google.com/view/lfdinelectrocautery}{https://sites.google.com/view/lfdinelectrocautery}
    
\end{abstract}

\section{Introduction}\label{sec:intro}
\input{intro}

\section{Related Work}\label{sec:related_work}
\input{related_work}

\section{problem definition}
\input{problem-definition}

\section{Method}\label{sec:method}
\input{method}

\section{Experiments}\label{sec:experiments}
\input{experiment}

\section{conclusion}\label{sec:conclusion}
\input{conclusion}

\bibliography{sample}

\end{document}

%% file: intro.tex
As medical care demands increase worldwide, the human surgeon shortage is becoming more pressing~\cite{zhang2020physician}. Training surgical robots for specific tasks has the potential to help decrease surgeon workload and enhance the precision of surgeries~\cite{attanasio2021autonomy}. However, surgical tasks are challenging as they require sequential decision making with complex deformable physical interactions and a low tolerance for error. Furthermore, an ideal surgical robot should be able to infer a human's underlying task objectives and intent even if optimal human demonstrations are not available~\cite{ginesi2020autonomous}. Our proposed approach uses pairwise preference labels over suboptimal trajectory data to capture the demonstrator's intent in the form of a learned reward function that can be optimized via reinforcement learning to yield a robust robot policy. \par

\begin{figure}[t]
    \centering
    \includegraphics[width=1.\columnwidth]{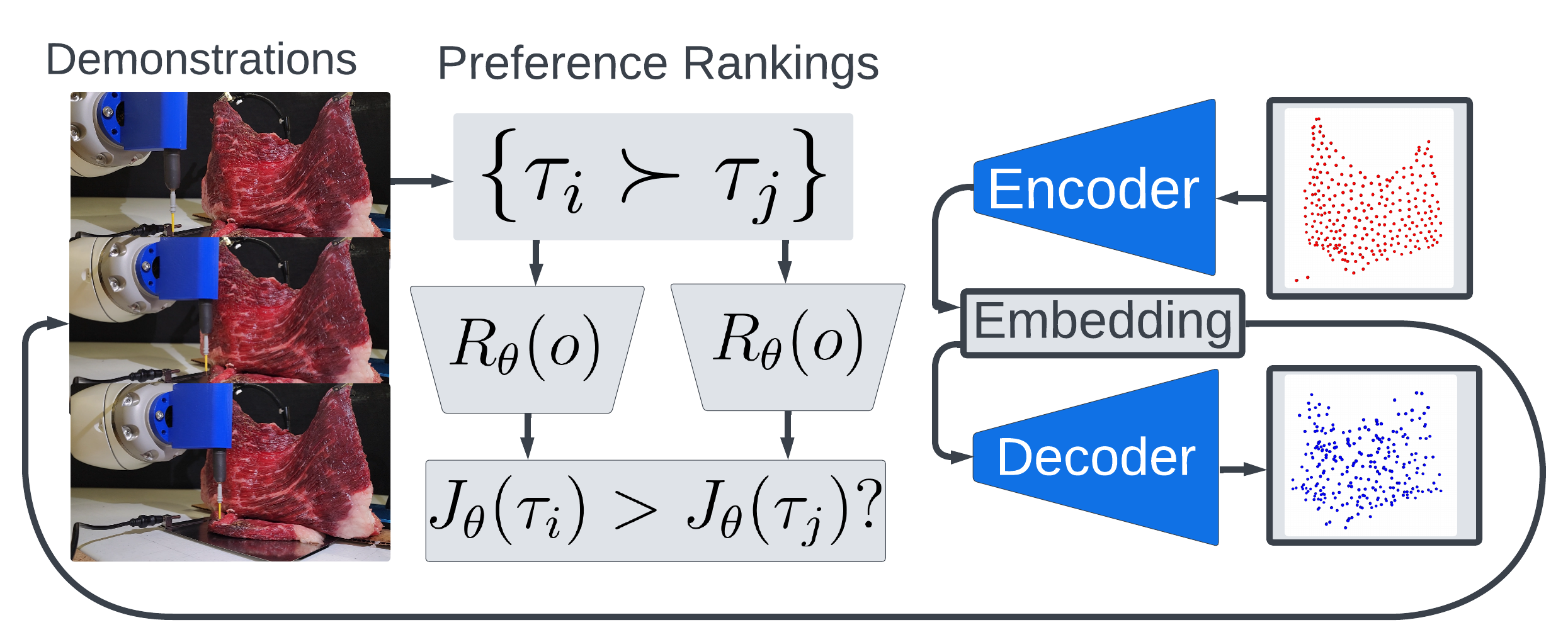}
    \caption{Our proposed method first learns a latent feature representation by pre-training an autoencoder to reconstruct partial-view point clouds. Then, given pairwise preferences over demonstrations with observations encoded by the latent feature representation, our method learns a reward function that maximizes the likelihood of the pairwise preferences.}
    \label{fig:method_summary}
    \vspace{-5pt}
\end{figure}

Our work builds on prior research on Learning from Demonstration (LfD), which has shown to be one of the most effective solutions for enabling robots to learn to perform complex tasks~\cite{argall2009survey,ravichandar2020recent,arora2021survey}.
However, existing methods often suffer from two major drawbacks. First is the assumption of fully observable states. While some past surgical LfD work has assumed full knowledge of object positions and properties~\cite {pore2021learning,huang2021toward, su2021toward}, in practice, a fully observable state space is not achievable for surgical robots with point cloud observations~\cite{kim2020autonomously,pore2021learning}. Second, most prior work in surgical LfD only considers optimal or near-optimal demonstrations~\cite{schwaner2021autonomous,pore2021learning}, which may not always be available. This assumption can lead to potential overfitting to the suboptimalities in the demonstrations and poor performance.~\cite{argall2009survey, ravichandar2020recent}. 

To address the problem of partial observability, we use a point cloud autoencoder to learn a low-dimensional feature vector of the partial point cloud scene. To address the problem of suboptimal demonstrations, we leverage ideas from prior work on learning reward functions based on preference labels over suboptimal trajectories~\cite{brown2019extrapolating} to learn a robust reward function suitable for surgical tasks with point cloud observation embeddings as input to the reward function. 

We first demonstrate our approach in two simulated surgical electrocautery tasks where we demonstrate 64.13\% and 44.70\% improvements over pure imitation learning, respectively. Next, we demonstrate proof of concept on a physical electrocautery task with ex vivo bovine muscle tissue, achieving five successful trials out of seven trials. Our work takes the first steps towards learning complex surgical tasks via reward learning from human feedback. Importantly, our approach is able to learn from preference labels over suboptimal task executions. This reduces the need for near-optimal demonstrations and opens the door to surgical policy learning from qualitative human evaluations.

%% file: related_work.tex
A variety of LfD approaches have been developed to solve tasks in the surgical domain. Kim et al. use behavioral cloning with image observation to automate tool navigation in retinal surgery~\cite{kim2020autonomously}. Huang et al. develop a policy network to automate context-dependent surgical tasks~\cite{huang2021toward}. Pore et al. combine generative adversarial imitation learning and model-free reinforcement learning (RL) to automate soft-tissue retraction~\cite{pore2021learning}. 
However, prior work requires near-optimal demonstrations. By contrast, we learn policies from suboptimal training data.  Furthermore, performing RL on the learned reward allows us to avoid the common LfD problem of compounding error since the learner visits states induced by its policy during training~\cite{swamy2021moments}. 
In contrast to prior work that leverages deep RL on surgical tasks~\cite{thananjeyan2017multilateral,pore2021safe, chiu2021bimanual}, our approach does not require a hand-crafted reward function and works with complex, partial observations of the surgical scene. \par

Our work is an instance of reinforcement learning from human feedback (RLHF)~\cite{wirth2017survey,christiano2017deep,ouyang2022training,tien2022causal}. Compared to near-optimal demonstrations and numerical reward labels, RLHF methods only require relative judgment over behaviors which is easier to provide~\cite{wirth2017survey,shin2023benchmarks}. Prior work leverages online human preferences over trajectories generated by an RL agent to update a learned reward function and a policy interactively~\cite{christiano2017deep}. However, on-policy RL is sample inefficient, so querying humans during policy learning may require a prohibitive amount of human supervision. Our research method is inspired by prior work on offline preference-based reward learning that leverages suboptimal demonstrations and learns a reward model from pairwise preferences over these trajectories, enabling better-than-demonstrator performance~\cite{brown2019extrapolating,brown2020better}.

Our work seeks to learn electrocautery robot policies from demonstrations. While electrocautery is a common surgical task~\cite{cordero2015electrosurgical,groot1994electrocautery,un2013systemic,morris2009electrosurgery,ismail2017cutting}, we are not aware of prior work applying learning from demonstrations and reward learning to electrocautery. We model surgical electrocautery as sequentially reaching attachment points between surfaces to remove them. Our work is similar to Krishnan et al.~\cite{krishnan2019swirl}, who approximate a long-horizon sequential task as a sequence of sub-tasks each represented by a local reward function learned from Inverse Reinforcement Learning (IRL)~\cite{arora2021survey}; however, prior work assumes demonstrations are optimal and execute sub-tasks in the same order for computational tractability. By contrast, we drop the restriction that the trajectory has to reach attachment points in a specific order, which enables learning from suboptimal demonstrations efficiently. \par

%% file: problem-definition.tex
We model our problem as a Partially-Observable Markov Decision Process (POMDP), which is formulated as a tuple of state space $\mathcal{S}$, action space $\mathcal{A}$, transition probability $\mathcal{T}$: $\mathcal{S}$ $\times$ $\mathcal{A}$ $\rightarrow$ $[0,1]$, observation space $\Omega$, emission probability $\mathcal{O}$: $\mathcal{S}$ $\rightarrow$ $[0,1]$, reward function $R$: $\mathcal{S}$ $\rightarrow$ $\mathbb{R}$, discount factor $\gamma$ $\in$ $[0,1]$ and horizon T~\cite{spaan2012partially}. We assume no direct access to the true reward function, nor the true state. Thus, we seek to learn the function $R_{\theta}$: $\Omega$ $\rightarrow$ $\mathbb{R}$ that approximates the actual reward function and leads to optimal behavior under the true, unobserved reward function. In the surgical robotics domain that we consider, $\Omega$ is the space of partial-view point clouds $\mathcal{P}$ $\subseteq$ $\mathbb{R}^{3}$ of the workspace scene augmented with the task-related robot state $s^{robot}$ such as end-effector position. A demonstration is defined as a trajectory $\tau$ = $(o_1, o_2, ... , o_{T})$ consisting of T observations. 
Note that $\tau$ could potentially be a suboptimal demonstration.
We denote that a trajectory $\tau_i$ is more preferred than $\tau_j$ by $\tau_j \prec \tau_i$. To learn the reward function, we assume access to a dataset of trajectories,  $D = \{\tau_i\}_{i=1}^M$, and access to pairwise preference rankings, $ \{(\tau_{i}, \tau_{j}, \mathds{1}_{\{ \tau_{i} \prec \tau_{j} \}}) :  \tau_{i}, \tau_{j} \sim D\}$.

%% file: method.tex
\begin{figure}[t]
  \centering
  \includegraphics[width=\columnwidth]{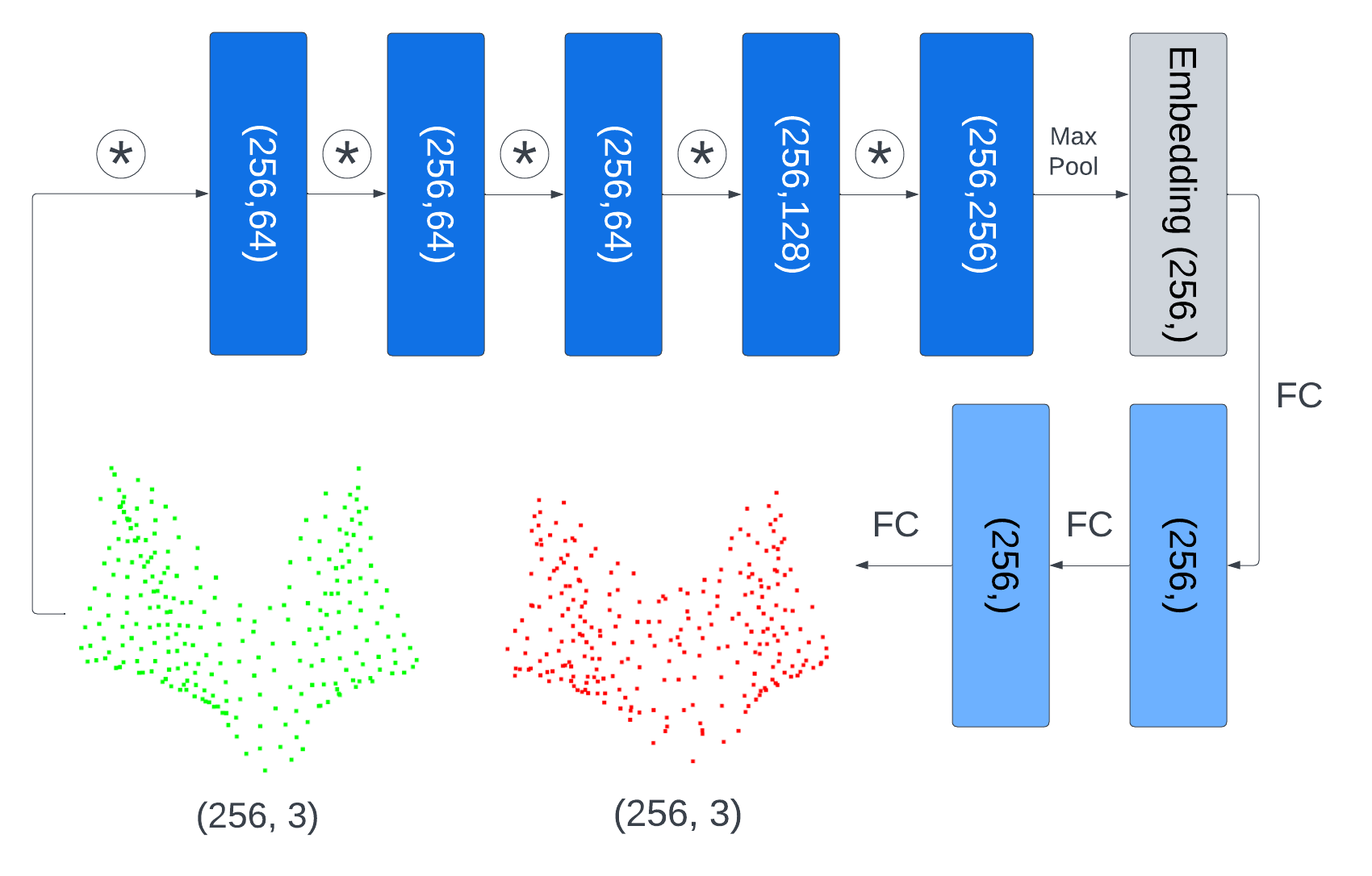}  
  \caption{Our autoencoder takes in the green point cloud and outputs the red reconstructed point cloud. * denotes (RELU $\circ$ group norm $\circ$ 1D convolution), FC denotes (ReLU $\circ$ linear layer) and the tuples denote the shape of the input to each layer. Convolution and group norm are done along the second dimension of the input. Max pooling is done along the first dimension of the input}
  \label{fig:autoencoder}
\end{figure}

As summarized in Fig. ~\ref{fig:method_summary}, we train an autoencoder to obtain low-dimensional feature representations of partial-view point clouds. These representations are then used to learn a parameterized reward function from preference rankings over trajectories of observations, and the learned reward function is used to train a policy. We discuss these steps below.

\subsection{Point Cloud Autoencoder}
Rather than directly training a reward model from partial-view point clouds, we suggest a more scalable approach that leverages a pre-trained point cloud autoencoder to map the high-dimensional point clouds into a lower-dimensional latent feature representation. We use this low-dimensional latent representation together with the task-related robot state as the input for the learned reward function.
Fig.~\ref{fig:autoencoder} shows the architecture of our point cloud autoencoder. We first downsample the point cloud to 256 points as a pre-processing step. We then pass the partial-view point cloud $P_{I}$ $\in$ $\mathbb{R}^{256 \times 3}$ through the encoder $\phi$: $\mathcal{P}$ $\rightarrow$ $\mathbb{R}^{256}$ consisting of five 1D convolution layers with non-linearity to get a latent feature vector with dimension 256. 
Using a fully-connected decoder $\psi$: $\mathbb{R}^{256}$ $\rightarrow$ $\mathcal{P}$, we decode this latent representation back to a reconstructed output point cloud $P_{o}$. Our reconstruction loss function is defined as $L = CD(P_{I}, P_{o}) + \lambda * EMD(P_{I}, P_{o})$, where the Chamfer Distance (CD)~\cite{wu2021density} is the sum of the squared distance of every point to the nearest point the other point cloud, and the Earth Mover's Distance (EMD)~\cite{wu2021balanced} computes the distance between distributions of point clouds by computing the minimum amount of work to transform one point set to another one. CD encourages matching the coarse geometry of the point clouds but not the density distributions of point clouds ~\cite{wu2021density}. A linear combination of CD and EMD encourages matching both large-scale and local geometry of point clouds. We explored several values for the tradeoff constant $\lambda$ and observed that best results occur when the initial $\lambda * EMD(P_{I}, P_{o})$ loss is approximately equal to one-fifth of the initial $CD(P_{I}, P_{o})$ loss. 

\vspace{-10pt}
\begin{equation} \nonumber
    \begin{aligned}
    &CD(P_{I}, P_{o}) = \sum_{x \in P_{I}} \min_{y \in P_{o}} ||x-y||_2^2 + \sum_{y \in P_{o}} \min_{x \in P_{I}} ||x-y||_2^2 \\
     &EMD(P_{I}, P_{o}) = \min_{B: P_{I} \rightarrow P_{o}} \sum_{x \in P_{I}} ||x - B(x)||_2
    \end{aligned}
\end{equation}
\vspace{-10pt}

\subsection{Preference-Based Reward Learning}

We assume a discount factor of $\gamma=1$ since trajectories have finite horizon T. We also assume that the preferences $\tau_i$ $\prec$ $\tau_j$ positively correlate with $J(\tau_i)<J(\tau_j)$ where $J(\tau) = \sum_{o \in \tau} R(o)$ is the return of a trajectory under the unobserved true reward function. We use Trajectory-ranked Reward Extrapolation (T-REX)~\cite{brown2019extrapolating} to learn a reward function that explains the pairwise preferences over demonstrations and potentially recovers the true reward function. Denote $R_{\theta}(o)$ as the parameterized learned reward function and define $J_{\theta}(\tau) = \sum_{o \in \tau} R_{\theta}(o)$ as the return of a trajectory $\tau$ according to learned reward function. 
The ideal learned reward function should satisfy the constraint $\forall \tau_i, \tau_j \sim \Pi, \tau_i \prec \tau_j \rightarrow J_\theta(\tau_i)<J_\theta(\tau_j)$, where $\Pi$ is the demonstration distribution. Thus we minimize the following loss function to learn the reward function $R_{\theta}(o)$ that maximizes the likelihood of the preference rankings: 
\vspace{-2pt}
\begin{equation} \label{eq:loss}
    L(\theta) = E_{\tau_i,\tau_j \sim \Pi}[\xi(P(J_\theta(\tau_i)<J_\theta(\tau_j)), \tau_i \prec \tau_j)]
\end{equation}
\vspace{-2pt}
where $\xi$ is cross-entropy loss and $P$ is a softmax-normalized probability distribution defined as follows:

\vspace{-10pt}
\begin{equation} \nonumber
    P(J_\theta(\tau_i)<J_\theta(\tau_j))\approx \frac{\exp \sum_{o \in \tau_j} R_\theta(o)}{\exp \sum_{o \in \tau_i} R_\theta(o)+\exp \sum_{o \in \tau_j}R_\theta(o)}
\end{equation}
\vspace{-10pt}

\vspace{-10pt}
\begin{equation} \nonumber
     L(\theta) = - \sum_{\tau_i \prec \tau_j} log \frac{\exp \sum_{o \in \tau_j} R_\theta(o)}{\exp \sum_{o \in \tau_i} R_\theta(o)+\exp \sum_{o \in \tau_j} R_\theta(o)}.
\end{equation}
\vspace{-10pt}

There are several potential ways to obtain trajectories and pairwise preference rankings in practice: (1) the trajectories can come from one or more non-expert human demonstrations, (2) they can be automatically generated by the robot~\cite{liu2023efficient} and then used as active queries for the human to compare, and (3) pairwise preferences can be automatically generated by adding noise to an imitation policy~\cite{brown2020better,liu2023efficient}. Notably, it has been shown that untrained individuals can generally assess surgical skills rapidly, efficiently, and accurately across different specialties and types of surgeries by watching surgical recordings~\cite{olsen2022crowdsourced}. Untrained individuals can rapidly provide evaluations on basic robotic surgical dry-laboratory tasks that highly correlate with expert evaluations~\cite{white2015crowd}. Given the learned reward function, a stochastic policy $\pi$: $\Omega$ $\times$ $\mathcal{A}$ $\rightarrow$ $[0,1]$ can be learned by maximizing the expected return $E[\sum_{t=1}^{T} \gamma^{t-1} R_{\theta}(o_t) | \pi]$ using any RL algorithm. Our approach is summarized in Algorithm \ref{alg:pref_learn_obs}. \par

\begin{algorithm}[t]
\caption{Preference-based Reward Learning with Partial Observations}\label{alg:pref_learn_obs}
1. Collect a random set of partial-view point clouds $D_{AE} = \{p_{i}\}_{i=1}^N$\\
2. Pre-train autoencoder on $D_{AE}$ with $\phi$: $\mathcal{P}$ $\rightarrow$ $\mathbb{R}^{256}$ as the encoder\\
3. Collect random demonstrations consisting of partial-view point cloud embedding concatenated with the task-related robot state\\ $D = \{\{(\phi(p_{t}),s_{t}^{robot})\}_{t=1}^T\}_{i=1}^M$\\
4. Collect pairwise preference rankings $D_{rank} = \{(\tau_{i}, \tau_{j}, \mathds{1}_{\{ \tau_{i} \prec \tau_{j} \}}) :  \tau_{i}, \tau_{j} \sim D\}$\\
5. Optimize $R_{\theta}$ by minimizing $L(\theta)$  on $D_{rank}$ (Eq.~\eqref{eq:loss})\\
6. Find the optimal policy $\pi$ under $R_{\theta}$ using RL\\
\end{algorithm}

\section{Policy Learning}
The observation space contains the task-related robot state $s^{robot}$ so that the robot gets dense reward at every action. We choose the action to be end-effector position $p_{eef} \in \mathbb{R}^3$ instead of end-effector velocity $v_{eef} \in \mathbb{R}^3$ or joint velocity $v_{joint} \in \mathbb{R}^k$ (k is the number of joints). \par

The benefit of such a design choice is two-fold. First, using $p_{eef}$ as the action makes RL more sample efficient. The task-related robot state added to the observation space should match the robot action. For example, if the action is $v_{joint}$, then the joint position should be added to the observation space instead of end-effector position. Otherwise, vastly different joint actions can result in the same task-related robot state, which makes learning the policy difficult. Hence, using $p_{eef}$ as the action keeps the dimension of the task-related robot state low, making RL more sample efficient. This action definition also causes larger end-effector dispalcement that may enable more efficient exploration. Second, using $p_{eef}$ as the action is more interpretable, as a trajectory in end-effector position explicitly indicates where the end-effector will reach in the next time step. \par

In order to command the robot in parallel in the RL environment, we transform end-effector actions output by the policy into joint velocity actions via a resolved rate controller~\cite{whitney1969resolved}. \par

%% file: experiment.tex
\begin{figure}[t]
     \centering
     \begin{subfigure}[b]{0.41\columnwidth}
         \centering
         \includegraphics[width=\textwidth]{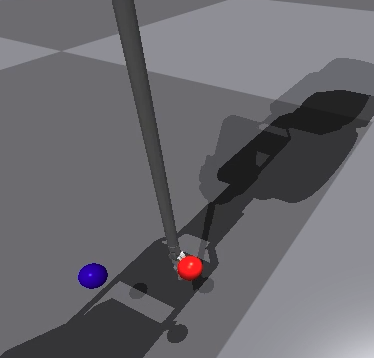}
         \caption{Sphere Task}
         \label{fig:sphere}
     \end{subfigure}
     \hspace{10pt}
     \begin{subfigure}[b]{0.45\columnwidth}
         \centering
         \includegraphics[width=\textwidth]{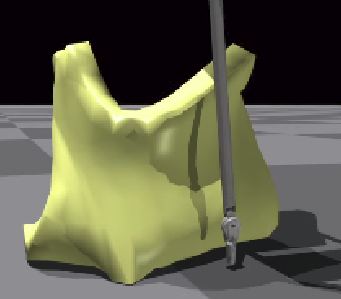}
         \caption{Cutting Task}
         \label{fig:cutting}
     \end{subfigure}
        \caption{Experimental setups with the dVRK surgical robot in the Isaac Gym simulator.}
        \label{fig:init_pic_cap}
\end{figure}

\subsection{Simulation Experimental Setup}

We first apply our method to simulated surgical electrocautery tasks. Simulated experiments are conducted in the Isaac Gym simulation environment \cite{liang2018gpu}, using a simulated patient-side manipulator of the da Vinci Research Kit (dVRK)~\cite{kazanzides2014open} robot. Point clouds of the workspace scenes are obtained via a simulated RGBD camera at a fixed position.

The surgical electrocautery task is modeled as sequentially moving the end-effector to attachment points between tissues and removing them. To simplify training, we assume having the end-effector reach the points is sufficient to remove them due to the limitations of the simulator (although we demonstrate real electrocautery in the physical experiments). \par

In the first simulated experiment (Sphere Task), the attachment points are represented as two spheres, and no tissues are present, as shown in Fig. ~\ref{fig:sphere}. The robot aims to move its end-effector to reach both spheres in any order. The reward function and the policy are trained on an augmented observation space, which concatenates the robot's end-effector's cartesian coordinates and the partial point cloud embedding of the sphere(s) output by the encoder. To simulate electrocautery, the sphere disappears when the end-effector reaches one sphere. Since the learned reward function depends on the observation, a change in the point cloud embedding provides a signal that the end-effector should reach the remaining sphere. \par

In the second simulation experiment (Cutting Task), we attach a simulated rectangular tissue onto a flat surface through a single attachment point sampled randomly, as shown in Fig. ~\ref{fig:cutting}. In order to reveal the attachment point, the tissue is retracted using a deterministic policy. The reward function and the policy are trained on an augmented observation space which is a concatenation of the robot's end-effector's cartesian coordinates and the partial point cloud embedding of the retracted tissue output by the encoder. The goal of the robot is to move its end-effector to the attachment point. The position of the attachment point must be inferred from the deformation of the retracted tissue. \par

\begin{table}[t]
\captionof{table}{Training data size, where K is the number of trajectory sets, C is the number of trajectories per set, and M is the number of preference rankings sampled with replacement.}\label{numDataTable}
\begin{center}
\begin{tabular}{ |c|c|c|c| }
 \hline
 Task & K & C & M \\ 
 \hline
 Sphere & 30 & 30 & 14000 \\ 
 Cutting & 60 & 30 & 14000 \\ 
 \hline
\end{tabular}

\vspace{-10pt}
\end{center}
\end{table}

 \subsection{Data collection}
To simulate suboptimal demonstrations, we programmatically collect a set of trajectories via a suboptimal motion planner. We generated multiple sets of random trajectories, each corresponding to a random configuration of the scene that the robot observes. The training data details, including the number of trajectory sets (K), the number of trajectories per set (C), and the number of preference rankings (M), are summarized in Table \ref{numDataTable}. Note that given $m$ ranked trajectories, we obtain $(m^2 - m)/2$ pairwise preferences. This allows us to obtain a large $M$ from a much smaller number of ranked trajectories. For the Sphere Task, a random configuration of the scene is the random cartesian coordinates of the two spheres. Spheres are sampled along a random horizontal straight line in the 3D space with slope in [-1,1]. For the Cutting Task, a random configuration of the scene is the retracted tissue with a random attachment point. Attachment points are sampled within a (2.5\,cm x 5\,cm) rectangle in the front half of the (20\,cm x 20\,cm) tissue closer to the robot. Trajectories are within the robot's workspace, which is a bounding box in the 3D space.\par

The random trajectories are generated as follows: given a fixed trajectory length and initial end-effector position, we sample the number of attachment point(s) to be reached. At each discrete timestep of the trajectory, we sample which attachment point should be reached. If no attachment point should be reached at this timestep, we sample a random point in the workspace to be reached. Finally, we execute the trajectory to collect a sequence of observations using inverse kinematics to control the end-effector. \par

For our experiments, we designed a ground-truth (GT) reward for ranking demonstrations to allow us to quantitatively measure how well our method recovers the ideal reward function and to allow us to better compare against baseline approaches. Note that our algorithm never observes the ground truth reward. To ensure the comparability of trajectory pairs using the GT reward function, only trajectories that have the same initial configuration of the scene are paired and ranked. Trajectory pairs are sampled randomly with replacement. The GT reward function is $$ R(eef, B) = \max_{b \in B} \frac{1}{|| eef-b||^2_2 + \epsilon}$$ where $eef$ is the 3D cartesian coordinates of the end-effector, $b$ is the 3D cartesian coordinates of an attachment point, $B$ is the set of attachment points and $\epsilon$ is a small number. We use $\epsilon=1e-4$ for Sphere Task and $\epsilon=1e-5$ for Cutting Task. \par

\begin{figure*}[tp]
     \centering
     \begin{subfigure}{0.30\textwidth}
         \centering
         \includegraphics[width=\linewidth]{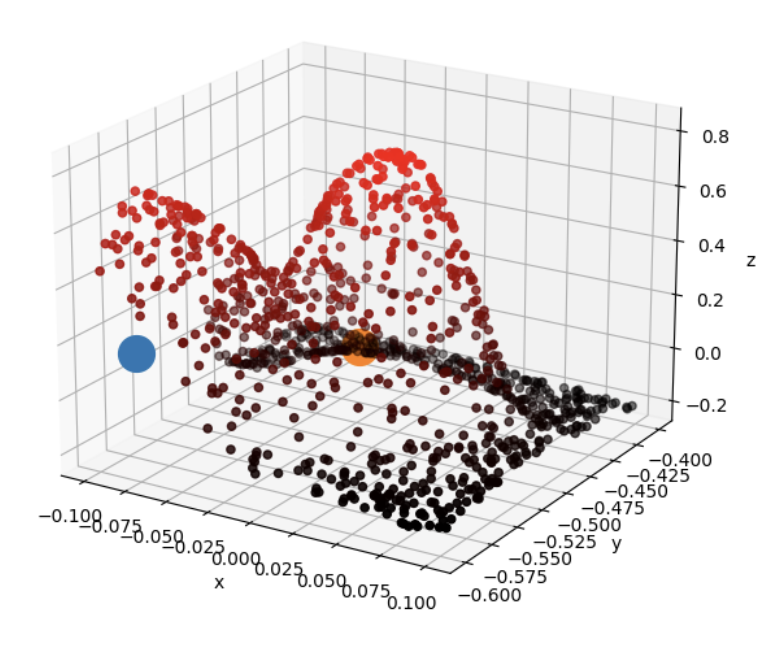}
         \caption{Sphere Task: heat map of learned reward when two spheres remain} 
         \label{fig:pred_reward_two}
     \end{subfigure}
     \hfill
     \begin{subfigure}{0.33\textwidth}
         \centering
         \includegraphics[width=\linewidth]{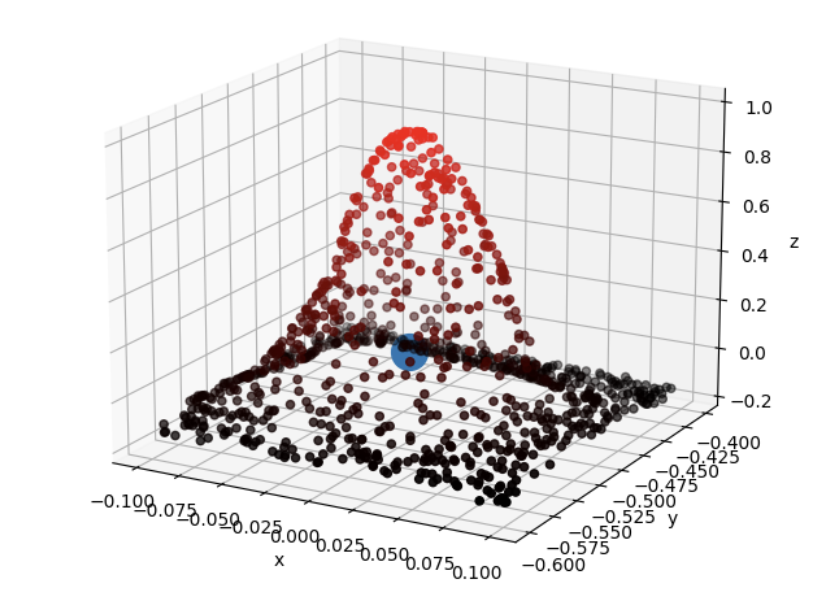}
         \caption{Sphere Task: heat map of learned reward when one sphere remains} 
         \label{fig:pred_reward_one}
     \end{subfigure}
     \hfill
     \begin{subfigure}{0.26\textwidth} 
         \centering
         \includegraphics[width=\linewidth]{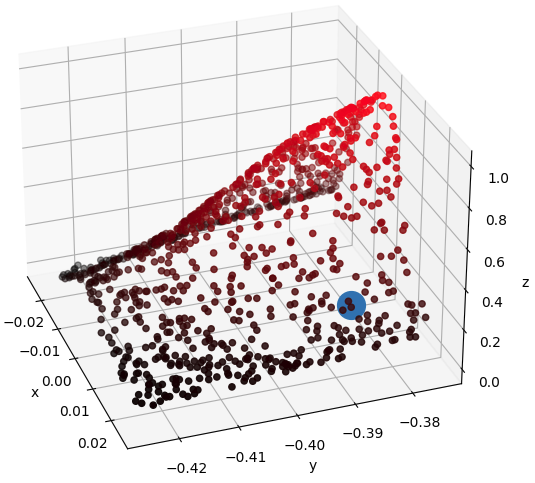}
         \caption{Cutting Task: heat map of learned reward} 
         \label{fig:pred_reward_cut_14000}
     \end{subfigure}
     \caption{Visualization: given a specified number of attachment point(s) in the scene, end-effector x-y positions (red points) are sampled on the same horizontal plane of the attachment point(s). The z-value of each red point is the predicted reward given the partial point cloud observation and the coordinates of the corresponding end-effector. Brighter color means higher predicted reward. }
     \label{fig:reward_heatmap}
\end{figure*}

We collect partial point clouds of random scene configurations in simulation to pre-train the autoencoder. For the Sphere Task, partial point clouds of 71,000 random positions of two spheres were collected. For each random position of the spheres, the data collection is repeated for different permutations of spheres disappearing. 
For the Cutting Task, partial point clouds are collected for 10,000 configurations of tissues each determined by a random attachment point. Since complex geometry can be reconstructed with just 10,000 random point clouds, the number of partial point clouds for the Sphere Task can be potentially lowered in the future. \par

\subsection{Policy Learning}
The policy is trained using Proximal Policy
Optimization (PPO)~\cite{schulman2017proximal}. We run 450 robots in parallel for efficiency purposes. Our action space is the space of end-effector position so that RL can be more sample efficient and explainable, so policies can be transferred between robots with different embodiments, and to enable easier sim2real transfer. In order to restrict the robot's cartesian action within the workspace, we use a sigmoid function $\sigma$ to clip the cartesian action along each dimension $i$ of {x,y,z} as follows:
$action \leftarrow min_i + (max_i - min_i)\sigma(action)$ where $max_i$ and $min_i$ are the upper and lower bounds of dimension i. \par

\begin{figure}[t]
     \centering
     \begin{subfigure}[thb]{0.9\columnwidth}
         \centering
         \includegraphics[width=\linewidth]{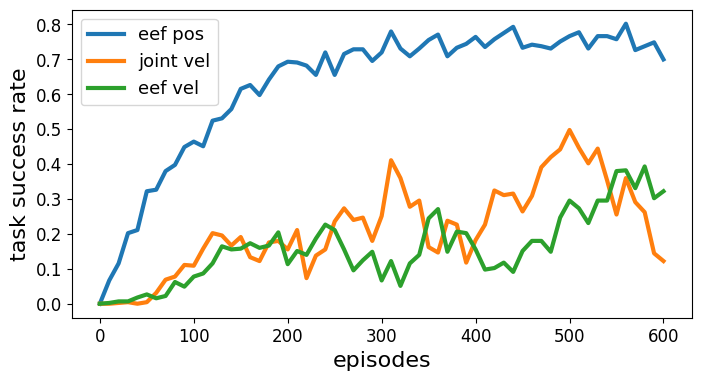}
     \end{subfigure}
     \caption{Learning curves of RL with different action spaces: (blue) end-effector position control, (green) end-effector velocity control, and (orange) joint velocity control. End-effector position control achieves the highest success rate.}
     \label{fig:action_compare}
\end{figure}

\begin{table}[t]
\captionof{table}{Testing accuracy of the learned reward function trained on decreasing numbers of pairwise preference rankings (from 13050 to 407) sampled without replacement}\label{SphereSampleEffTable}
\begin{center}
\begin{tabular}{ |c|c|c|c|c|c|c| }
 \hline
 Task & 13050 & 6525 & 3262 & 1631 & 815 & 407 \\ 
 \hline
 Sphere & 0.867 & 0.866 & 0.859 & 0.81 & 0.766 & 0.721 \\ 
 Cutting & 0.769 & 0.779 & 0.794 & 0.774 & 0.764 & 0.794 \\ 
 \hline
\end{tabular}

\end{center}
\end{table}

\begin{figure}[th!]
     \centering
     \begin{subfigure}{1\linewidth}
         \centering
         \includegraphics[width=\linewidth]{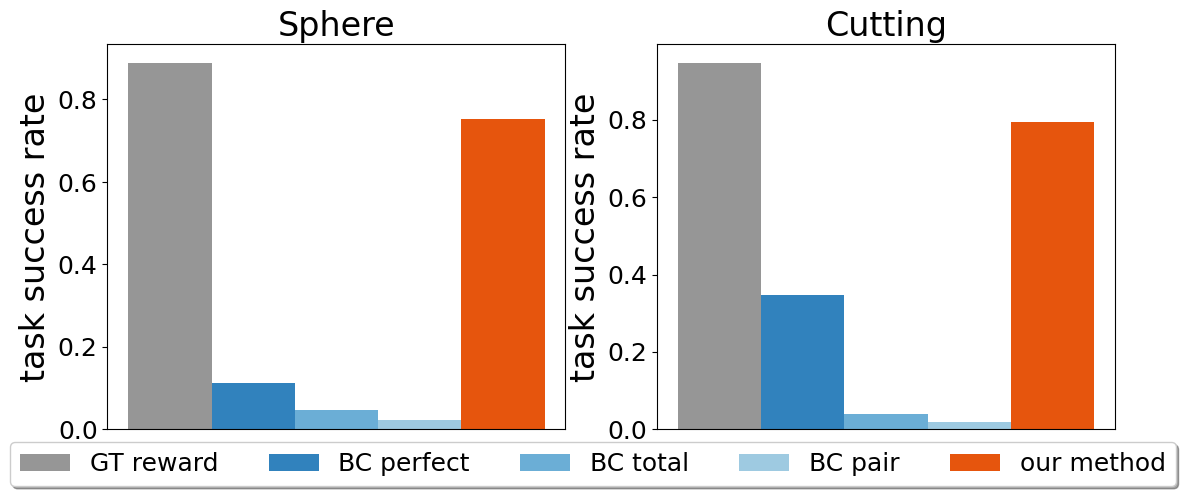}
         \label{fig:baseline_1}
     \end{subfigure}
     \caption{Baseline comparison in terms of task success rate for both tasks. Our method achieves success rates close to those of the GT reward oracle baseline. }
     \label{fig:baseline}
\end{figure}

\subsection{Simulation Experiment Results}

For both tasks, Fig.~\ref{fig:reward_heatmap} visualizes that our learned reward function captures the intention of the demonstrations well as it assigns high reward to observations where the end-effector position is close to any remaining attachment point(s). Fig.~\ref{fig:action_compare} also shows that commanding end-effector position control actions is important for maximizing RL performance. 

\begin{figure*}[ht]
     \centering
     \includegraphics[width=\linewidth]{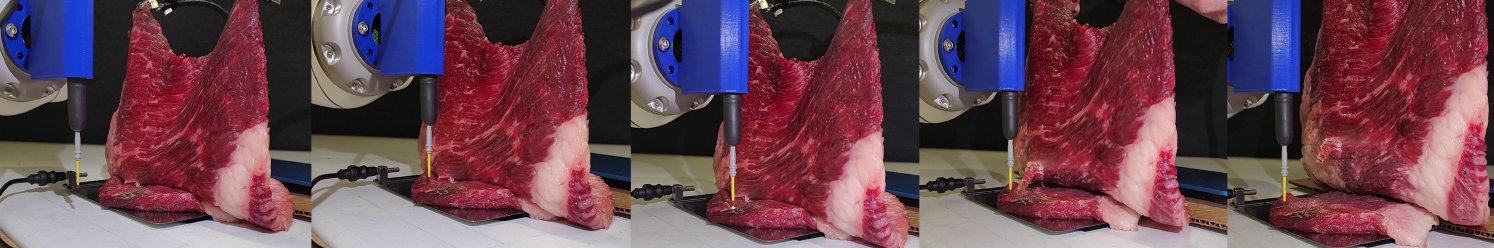}
     \caption{Sample of successful manipulation sequence in real robot experiment.} 
     \label{fig:exp_seq}
     
     \vspace{-10pt}
\end{figure*}

For both tasks, we compare against a Behavioral Cloning (BC) baseline~\cite{pomerleau1988alvinn}, a standard approach for learning from demonstrations~\cite{argall2009survey} that uses supervised learning to learn a policy that maps from states to actions. \textit{BC pair} denotes the policy trained on the more preferred demonstrations in every pairwise preference, \textit{BC total} denotes the policy trained on the top 20 percent of the suboptimal demonstrations, and \textit{BC perfect} denotes the policy trained on expert demonstrations of the same amount as suboptimal demonstrations.
For Sphere Task, Fig.~\ref{fig:baseline} shows that our method achieves close to 80 percent task success rate, upper-bounded by the 85 percent task success rate achieved by the policy trained on the ground-truth (GT) reward. For Cutting Task, 
Fig.~\ref{fig:baseline} shows that our method achieves 80 percent task success rate, upper-bounded by the almost 90 percent task success rate achieved by the policy trained on GT reward. 
 Fig.~\ref{fig:baseline} shows that our method outperforms BC policies that have access to the same amount of demonstrations: our method yields improvements of 64.13\% for Sphere Task and 44.70\% for Cutting Task over BC perfect.  \par 

To evaluate the sample efficiency of our method, we repeatedly halved the training data size and computed the testing accuracy of the learned reward function as shown in table~\ref{SphereSampleEffTable}. 
We empirically found that 6,525 pairwise preferences are needed to learn a robust reward function that achieves a task success rate of 80 percent in Sphere Task after policy learning. 
For Cutting Task, we empirically found that only 815 pairwise preferences are needed to learn a robust reward function that achieves a task success rate of 80 percent. \par

\subsection{Real Robot Experimental Setup}

\begin{figure}[t]
    \centering
    \includegraphics[width= \columnwidth]{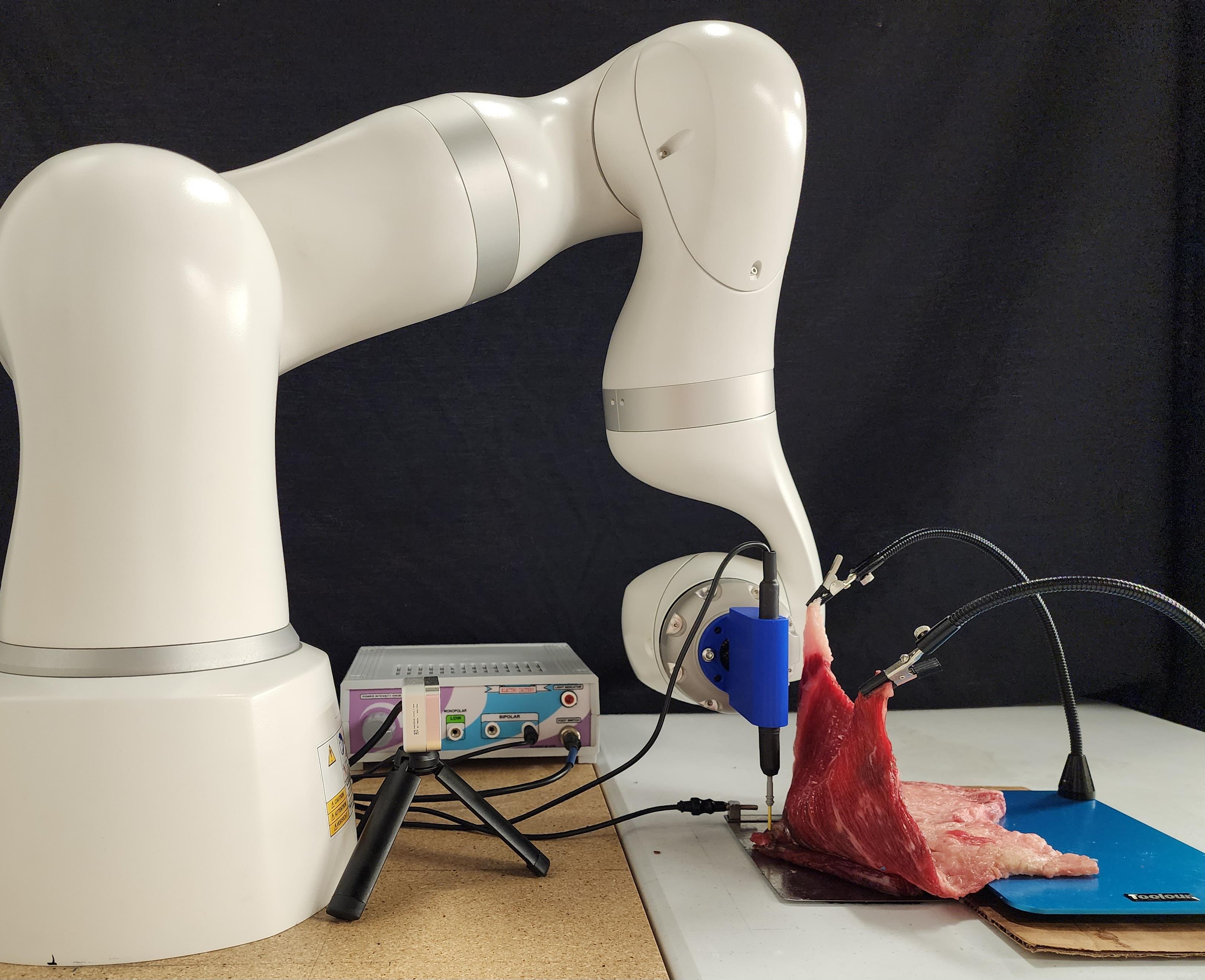}
    \caption{Experimental setup for  electrocautery cutting task}
    \label{fig:experiment_setup}
    \vspace{-10pt}
\end{figure}

A successful execution of the electrocautery policy is shown in Fig.~\ref{fig:exp_seq} and our full physical experimental setup is illustrated in Fig.~\ref{fig:experiment_setup}. We use bovine muscle tissue as an ex vivo tissue evaluation platform. We use one piece of tissue as a flat surface to which another retracted piece of tissue is attached. In each experimental trial, the retracted tissue is attached to a point of the flat surface within a 2.5\,cm x 5\,cm rectangle. An electrocautery tool is mounted on the end-effector of a KUKA LBR Med Robot. An Intel Realsencse depth camera D405 is employed in our setup for recording the tissue point cloud. \par

We test whether we can directly transfer the learned reward function, autoencoder, and policy trained in simulation to our real experimental setup. Given the fixed initial end-effector position and the initial point cloud of the scene containing the retracted tissue, we generate 200 end-effector trajectories, each of length 30, using the stochastic learned policy in an open-loop manner. The trajectory with the highest predicted learned reward is executed. 
We terminate the open-loop trajectory generated by the learned policy when the trajectory converges at the attachment point, and the policy switches to a heuristic cutting motion that oscillates left and right on the attachment point to remove it. To compute convergence of the open loop policy, at each timestep, we compute the component-wise mean and standard deviation of the end-effector positions from the initial timestep to the current timestep. This results in a vector of means and a vector of standard deviations at each timestep. As the end-effector motion from the learned policy converges to an attachment point, we expect the l2 norm of the difference of successive mean vectors and successive standard deviation vectors to decrease. The open-loop trajectory of the learned policy is terminated when these two numbers are lower than a threshold. We found that 0.005 as the threshold for the difference of mean vectors and 0.001 for the threshold on the standard deviation vectors worked well in practice.\par

\subsection{Real Robot Experiment Results}
We conducted seven trials of the experiment, each featuring a different attachment point location. The robot successfully accomplished both reaching the attachment points and executing the cutting task in 5 of the 7 trials.
In the other two experiments, the robot end-effector approached very close to the attachment points but ultimately halted prematurely before reaching the desired locations.
Upon careful analysis of these particular instances, we identified a common factor: the point clouds associated with these failure cases were out of distribution and consequently poorly reconstructed. This discrepancy in reconstruction adversely affected the quality of the latent embedding used for feature representation, resulting in a suboptimal policy. A plausible cause of this problem is the visual disparity between real tissue and the simulated tissue object used for training data collection.

%% file: conclusion.tex
In this paper, we propose a novel preference-based reinforcement learning approach that is well suited for partially observable surgical tasks. Our empricial results in simulation demonstrate that our approach is superior to pure imitation learning and is able to achieve high task success despite only having access to suboptimal demonstrations. We demonstrate that our method achieves 80\% task success rate in two simulated surgical electrocautery tasks. We also demonstrate a proof of concept physical surgical electrocautery task, in which our method achieved five successful trials out of seven total trials. 
Future research includes conducting a user study to evaluate how well non-expert humans rank demonstrations and how sensitive the learned reward is to noisy preference rankings. Since large numbers of offline demonstrations and preferences can be prohibitive, incorporating sample-efficient active reward learning~\cite{brown2018risk,biyik2019asking,shin2023benchmarks} into our approach is also an important future research direction. 